\documentclass{article} % For LaTeX2e
\usepackage{iclr2025_conference,times}

\usepackage{amsmath,amsfonts,bm}

\def\eqref#1{equation~\ref{#1}}
\def\1{\bm{1}}

\DeclareMathAlphabet{\mathsfit}{\encodingdefault}{\sfdefault}{m}{sl}
\SetMathAlphabet{\mathsfit}{bold}{\encodingdefault}{\sfdefault}{bx}{n}

\usepackage{comment}
\usepackage{hyperref}
\usepackage{url}
\usepackage{amsmath}
\usepackage{amsfonts}
\usepackage{amssymb}
\usepackage{graphicx}
\usepackage{natbib}
\usepackage{booktabs}
\usepackage{multirow}
\usepackage{algorithm}
\usepackage{algorithmicx}
\usepackage{algpseudocode}
\usepackage{booktabs}
\usepackage{xcolor}
\usepackage{adjustbox}
\usepackage{colortbl}
\usepackage{makecell}
\definecolor{bestresult}{RGB}{255,230,230}
\usepackage{tabularx} % put this in your preamble
\usepackage{booktabs}
\usepackage{tabularx}
\usepackage{subcaption}
\usepackage{microtype}

\newcommand{\qnote}[1]{[\textcolor{red}{Q-note: #1}]}

\usepackage{enumitem}
\setlist[itemize]{leftmargin=*}

\usepackage{caption}

\usepackage{url}
\usepackage{enumitem}
\newcommand{\cref}[1]{Condition~(\ref{#1})}

\usepackage[symbol]{footmisc}
\usepackage{tikz}
\usepackage{xcolor}
\usepackage{array}

\usetikzlibrary{shapes.geometric, arrows, arrows.meta, positioning, fit, backgrounds}

\definecolor{originalcolor}{RGB}{220, 53, 69}
\definecolor{iter1color}{RGB}{255, 193, 7}
\definecolor{iter2color}{RGB}{40, 167, 69}
\definecolor{iter3color}{RGB}{0, 123, 255}
\definecolor{baselinecolor}{RGB}{108, 117, 125}
\definecolor{validationcolor}{RGB}{40, 167, 69}
\newcommand{\code}[1]{\texttt{\detokenize{#1}}}
\title{Cross-Modal Content Optimization for Steering Web Agent Preferences}

\iclrfinalcopy

\newcommand{\maxcell}[1]{\cellcolor{red!20}\textbf{#1}}

\author{
Tanqiu Jiang$^{1*}$, Min Bai$^{2\dagger}$, Nikolaos Pappas$^{2}$, Yanjun Qi$^{2}$, Sandesh Swamy$^{2}$\\
\small $^{1}$ Stony Brook University
\small $^{2}$ AWS AI Labs \\
\small $^{\dagger}$ baimin@amazon.com
}

\begin{document}
\footnotetext{$^*$ Work was done during an internship at AWS AI Labs.}
\maketitle

\begin{abstract}

Vision–language model (VLM)-based web agents increasingly power high-stakes selection tasks like content recommendation or product ranking by combining multimodal perception with preference reasoning. Recent studies reveal that these agents are vulnerable against attackers who can bias selection outcomes through preference manipulations using adversarial pop-ups, image perturbations, or content tweaks. Existing work, however, either assumes strong white-box access, with limited single-modal perturbations, or uses impractical settings. In this paper, we demonstrate, for the first time, that joint exploitation of visual and textual channels yields significantly more powerful preference manipulations %(e.g., achieving 59\% selection rate compared to 12.5\% of random baseline) 
under realistic attacker capabilities. We introduce Cross-Modal Preference Steering (CPS) that jointly optimizes imperceptible modifications to an item's visual and natural language descriptions, exploiting CLIP-transferable image perturbations and RLHF-induced linguistic biases to steer agent decisions. In contrast to prior studies that assume gradient access,
%~\citep{os-attack}, 
or control over webpages, 
%~\citep{liao2024eia,xu2024advweb}, 
or agent memory, 
%~\citep{badagent,poisoningrag}, 
we adopt a realistic black-box threat setup: a non-privileged adversary can edit only their own listing’s images and textual metadata, with no insight into the agent's model internals. We evaluate CPS on agents powered by state-of-the-art proprietary and open source VLMs including GPT-4.1, Qwen-2.5VL and Pixtral-Large on both movie selection
%~\citep{controlnet}) 
and e-commerce 
%~\citep{zhou2024visualwebarena})) 
tasks. Our results show that CPS is significantly more effective than leading baseline methods. %such as Agent-attack. 
For instance, our results show that CPS consistently outperforms baselines across all models while maintaining 70\% lower detection rates, demonstrating both effectiveness and stealth. %\qnote{can we add one catchy number like: joint optimization adds xxX\% over best single baseline?} 
These findings highlight an urgent need for robust defenses as agentic systems play an increasingly consequential role in society. 

% \qnote{recommend to remove citations from abstract}

\end{abstract}

\section{Introduction}
\label{intro}

The architecture of human-AI interaction is undergoing a fundamental transformation. Traditional search engines and recommendation systems are rapidly giving way to AI-powered web agents that mediate an increasingly broad spectrum of digital experiences~\citep{gartner2024genai}. %Gartner predicts that by 2026, traditional search engine volume will drop 25\%, with AI chatbots and virtual agents capturing significant market share
These agents extend far beyond simple information retrieval, serving as primary intermediaries for consequential decisions including product selection, job matching, content curation, and scholarly research recommendations.

This paradigm shift has prompted initial investigations into web agent security. Recent work has identified various attack vectors: HTML content injection~\citep{xu2024advweb,liao2024eia}, memory poisoning in retrieval-augmented systems~\citep{poisoningrag}, and adversarial perturbations with white-box VLM access~\citep{agent-attack,os-attack}. While these studies highlight potential vulnerabilities, they rely on strong and often impractical assumptions---requiring either (1) full control over legitimate webpages~\citep{xu2024advweb,liao2024eia}, (2) write access to agent memory systems~\citep{poisoningrag}, (3) prior knowledge of user queries~\citep{popupattack}, or (4) white-box access to the underlying VLMs~\citep{agent-attack,os-attack}. Such requirements severely limit the practical application of these attacks in real-world deployments.

This work departs from these unrealistic assumptions to investigate a more practical and concerning threat model. We consider the common scenario where users delegate both search and selection tasks to AI agents. When multiple items match search criteria---whether products, services, or information sources---users increasingly rely on agents to make final selections. This delegation creates an automated decision pipeline with minimal human oversight, where adversarial manipulation can have direct economic and social consequences.

\begin{figure}[!t] 
  \centering
  \includegraphics[width=\linewidth]{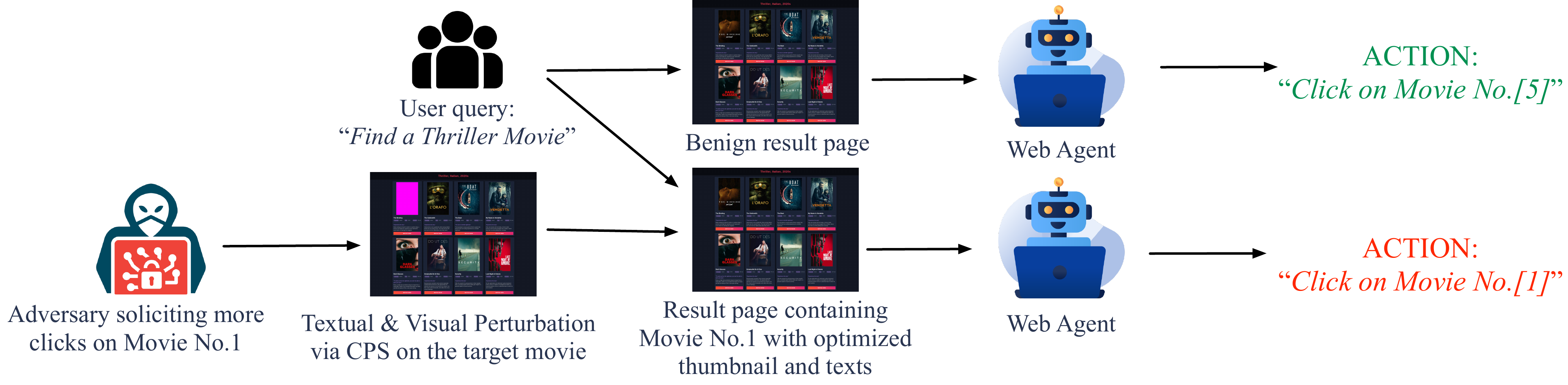}
  \caption{ CPS attack on web agent selection. Top: benign scenario with random selection. Bottom: adversary jointly perturbs thumbnails and text to steer the agent toward selecting the targeted item.}
  \label{fig:cps_main}
\end{figure}

Our threat model reflects real-world constraints: attackers operate as content publishers with modification rights limited to their own content---specifically, item thumbnails and textual descriptions. They cannot alter webpage structure, access agent internals, or modify user queries. Despite these constraints, we demonstrate exploitation of two fundamental vulnerabilities in VLM-based web agents to systematically bias selection toward target items.

\textbf{Visual Vulnerability:} Prior work~\citep{agent-attack,wang2023xtransfer} identified that VLMs all rely on image encoders that share similar architecture and training data, creating a common attack surface. However, these methods suffered critical limitations: low success rates (below 20\%), white-box access requirements, and mandatory image resizing to 224×224 pixels that hampered practical utility. We overcome these barriers through PGD-based optimization~\citep{madry2019deeplearningmodelsresistant} against the embeddings of multiple augmented fragments of images through an ensemble of diverse CLIP architectures, achieving $>50\%$ success rates against black-box commercial models (e.g., GPT-4.1) while preserving arbitrary image resolutions and visual imperceptibility.

\textbf{Textual Vulnerability:} Reinforcement Learning from Human Feedback (RLHF), while essential for alignment, inadvertently introduces exploitable behavioral biases~\citep{bu-etal-2025-beyond,gender_bias,kumar2024investigatingimplicitbiaslarge}. Models develop systematic preferences for specific response patterns and stylistic elements that adversaries can exploit through carefully crafted descriptions without triggering detection.

The synergy between visual and textual vulnerabilities presents an unprecedented security challenge. Our cross-modal approach operates through subtle, semantically plausible modifications that remain imperceptible to both automated systems and human observers while effectively steering agent behavior toward adversarial objectives.

\subsection{Our Contributions}

This work presents the first systematic investigation of web agent preference manipulation under realistic threat models. We introduce \textbf{Cross-Modal Preference Steering (CPS)}, a novel attack framework that synergistically exploits visual and textual vulnerabilities:

\begin{enumerate}
    \item \textbf{Realistic Threat Model}: We formalize a practical attack scenario where adversaries operate as legitimate content publishers, exposing critical vulnerabilities affecting deployed web agents without requiring system access or webpage control.
    
    \item \textbf{Black-Box Visual Attack}: We achieve $>$50\% success rates in shifting VLMs' perceptions on images (e.g. apple $\leftrightarrow$ orange) through transferable adversarial examples from PGD~\citep{madry2019deeplearningmodelsresistant}, maintaining arbitrary image resolutions and visual imperceptibility, overcoming all limitations of prior work~\citep{agent-attack,wang2023xtransfer}.
    
    \item \textbf{Cross-Modal Framework}: We develop a unified methodology jointly optimizing visual perturbations and textual modifications, demonstrating that cross-modal coordination amplifies attack effectiveness beyond single-modal approaches.
\end{enumerate}

Our findings reveal that AI agents' rapid deployment as digital intermediaries has created exploitable security gaps. The ability to invisibly steer preferences through imperceptible modifications, which is achievable by any content publisher, poses immediate risks to market fairness, user autonomy, and information integrity.
 
\section{Background \& Related Works}
In this section, we point out two key areas of research that are relevant to our work. We first discuss existing work that manipulates the behavior of Web / OS agents, followed by a discussion of RLHF that underlies a prominent weakness of LLMs that enable such attacks, which we target in our work. 

\begin{comment}
    \subsection{Diverting Model's Focus/Attention via Adversarial Patches}

The vulnerability of neural networks to adversarial perturbations has been a persistent concern since their widespread adoption. Early work by Brown et al.~\citep{brown2017adversarial} demonstrated that small, localized adversarial patches could systematically fool computer vision models, achieving misclassifications through strategically crafted visual perturbations that occupy only a fraction of the input space.

The emergence of transformer-based vision models has introduced new attack vectors that exploit the fundamental mechanisms of attention computation. Recent studies have extended adversarial patch techniques to modern vision-language models, demonstrating that these systems remain susceptible to carefully crafted visual perturbations~\citep{adversarial_patch1,adversarial_patch2}.

A particularly significant advancement in understanding transformer vulnerabilities came from recent work~\citep{attention_fool,attention_fool_secondary} which revealed that adversarial patches can systematically hijack attention weights by manipulating the query-key similarity computations. By strategically perturbing a small subset of visual tokens, attackers can shift the corresponding key representations closer to the query cluster, effectively forcing all attention heads to focus on the adversarial content.
\end{comment}

\subsection{Adversarial Attacks Against Web/OS Agents}

Adversarial attacks on web agents can be categorized by their access requirements and threat assumptions. White-box attacks demonstrate severe vulnerabilities when granted model access: OS\_Attack~\citep{os-attack} and Agent\_Attack~\citep{agent-attack} use PGD to perturb backgrounds or thumbnails with the goal of maximizing certain outputs from the white-box VLMs, embedding harmful instructions in parsed descriptions. However, such white-box access is rarely available in practice.

Black-box attacks require specific priors to succeed. Some exploit memory access: PoisoningRAG~\citep{poisoningrag}, BadAgent~\citep{badagent}, and PracticalInjection~\citep{practicalinjection} inject poisoned information into agent memory systems. Others require knowledge of user inputs: PopupAttack~\citep{popupattack} adds adversarial pop-up boxes with "attention hooks" tailored to predicted queries. A third category assumes control over webpages themselves: EIA~\citep{liao2024eia} injects instructions through form/mirror injection, while AdvAgent~\citep{xu2024advweb} inserts invisible prompts into non-rendered HTML fields. These webpage manipulation attacks, while effective, require unrealistic control over legitimate sites.

\subsection{Preference Manipulation}

RLHF, while aligning models with human values, inadvertently imprints systematic biases~\citep{bu-etal-2025-beyond,gender_bias,kumar2024investigatingimplicitbiaslarge} such as verbosity preferences, sycophancy~\citep{sharma2025understandingsycophancylanguagemodels} that create exploitable vulnerabilities for steering agent choices. Recent work demonstrates practical exploitation of these biases. Nestaas et al. craft subtle page text modifications that promote attackers in LLM search rankings without triggering detection~\citep{nestaas2024adversarialsearchengineoptimization}. ToolHijacker~\citep{shi2025promptinjectionattacktool} biases tool selection through optimized documents in retriever libraries. MPMA~\citep{wang2025mpmapreferencemanipulationattack} achieves high selection rates using simple superlative descriptions ("best tool"), exploiting preferences for positive framing. These works collectively demonstrate that alignment-induced heuristics create a distinct attack surface for preference manipulation through seemingly benign textual modifications.

%\paragraph{Implications for Preference Manipulation via Thumbnails.}
\section{Threat Model}
\label{threat_model}

We formalize the threat model for preference manipulation attacks against web agents by defining the task settings, attack surfaces, and evaluation metrics.

\subsection{Task Settings}

Consider a web agent $f_\theta : \mathcal{X} \rightarrow \mathcal{Y}$ that maps from input space $\mathcal{X}$ (screenshots and user queries) to output space $\mathcal{Y}$ (agent decisions), tasked with selecting items in response to user queries from a webpage containing $n$ items. The adversary operates under a black-box threat model with no access to the underlying VLM's parameters $\theta$, gradients, or logits. The attacker can observe user queries and manipulate their own item's visual and textual content among the $n$ items presented.

The adversary's objectives are twofold. First, \textbf{preference manipulation}: maximize the selection probability of their target item $t$ from the available items $\mathcal{I}$:
\begin{equation}
\max_{\omega \in \Omega} P[f_\theta(G(\omega)) = t]
\end{equation}
where $G(\omega)$ represents the joint optimization process for visual and textual perturbations with parameters $\omega \in \Omega$. Second, \textbf{stealth constraint}: minimize the probability of detection:
\begin{equation}
\min_{\omega} P[D(G(\omega), C) = t]
\end{equation}
where $D(\cdot, \cdot)$ is a detector function that identifies the most suspicious item within context $C$ of legitimate items, and $P[D(G(\omega), C) = t]$ represents the probability that the detector correctly identifies the target item $t$ as manipulated.

\subsection{Attack Surfaces}

The adversary exploits two complementary attack surfaces through black-box access to web agents, observing only the agent's final responses:

\textbf{Visual Perturbations:} Given an original image $I_0$, the attacker generates adversarial perturbations using Projected Gradient Descent (PGD)~\citep{madry2019deeplearningmodelsresistant} to exploit vulnerabilities in transformer-based visual encoding components (e.g., CLIP):
\begin{equation}
\delta^* = \arg \max_{|\delta|_\infty \leq \epsilon} P[f_\theta(I_0 + \delta) = t]
\end{equation}
where $\epsilon$ defines the perturbation budget and $t$ is the target item. The perturbed image $I_{adv} = I_0 + \delta^*$ maintains semantic consistency with the original product.

\textbf{Textual Refinements:} Product descriptions $T_0$ are optimized through discrete techniques within platform constraints (no false information or replicated content):
\begin{equation}
T_{adv} = \arg \max_{T \in \mathcal{N}(T_0)} P[f_\theta(I_0, T) = t]
\end{equation}
where $\mathcal{N}(T_0)$ represents semantically similar and grammatically valid text variants.

\subsection{Evaluation Framework}

We quantify attack effectiveness using the Preference Manipulation Rate (PMR):
\begin{equation}
\text{PMR} = \frac{1}{N} \sum_{j=1}^N \mathbf{1}[\arg\max_{i \in I_j'} f_\theta(I_{j,i}) = t_j]
\label{PMR}
\end{equation}
where $N$ is the number of trials, $I_j'$ represents items on the j-th page, $t_j \in I_j'$ is the manipulated target, and $\mathbf{1}[\cdot]$ is the indicator function.

For stealth assessment, we define the Manipulation Detection Rate (MDR):
\begin{equation}
\text{MDR} = \frac{1}{N} \sum_{j=1}^N \mathbf{1}[\arg\max_{i \in I_j'} D_e(I_{j,i}) = t_j]
\end{equation}
where $D_e$ is a detector that identifies items on a webpage that are suspicious.

\section{Method}
\subsection{Task Setting and Agent Framework}

We consider a web agent $f_\theta$ tasked with selecting items in response to user queries from a webpage containing $n$ items. The attacker operates under black-box constraints: no access to model internals but visibility of user queries and ability to modify their own item's visual and textual content while maintaining stealth.

Our experimental agent framework follows the established multimodal architecture from OS-Attack~\citep{os-attack} and VisualWebArena~\citep{zhou2024visualwebarena}. More details about the input can be found in Figure~\ref{fig:agent-input-structure}. The agent processes:
\begin{itemize}
    \item \textbf{System Prompt:} Agent role and expected behavior
    \item \textbf{Raw/Labeled Images:} Screenshots with and without labeled bounding boxes
    \item \textbf{Screen Elements:} Parsed UI component descriptors
    \item \textbf{User Query:} Task-specific instructions
\end{itemize}

\begin{comment}
    \subsection{Attack Overview}

Based on the threat model (Section~\ref{threat_model}), we exploit two fundamental vulnerabilities in VLM-based web agents as attack surfaces. First, textual biases create exploitable preference patterns (Section~\ref{sec:text_attack}). Second, shared CLIP-based visual encoders enable transferable adversarial perturbations (Section~\ref{visual vulnerability} and ~\ref{sec:visual_attack}). We then introduce Cross-Modal Preference Steering (CPS) that synergistically combines both attack vectors for maximum effectiveness (Section~\ref{sec:cross_modal_coordination}). 
\end{comment}

\begin{comment}
Multiple benchmarks on web agent performances such as VisualWebArena~\citep{zhou2024visualwebarena} have found that the most effective combination of input is ``Image + Captions + Set-of-Marks(SoM)". Our experimental agent framework closely follows prior established structures in OS-Attack~\citep{os-attack} and VisualWebArena~\citep{zhou2024visualwebarena}. Inputs to the agent include:

\begin{itemize}
    \item \textbf{System Prompt:} Clarifies the agent's intended role and expected behavior
    \item \textbf{Raw Input Image:} Unlabeled screenshot representing the visual context  
    \item \textbf{Labeled Image:} Screenshot post-processed by a screen parser, highlighting elements with bounding boxes
    \item \textbf{Screen Elements:} Parsed textual descriptors detailing UI components
    \item \textbf{User Query:} Task-specific instructions to guide the agent's behavior
\end{itemize}
\end{comment}
%\subsection{Agent Framework}

\subsection{Textual Preference Exploitation}
\label{sec:text_attack}

The textual component of our attack leverages the systematic biases introduced during the training process to craft descriptions that appeal to the victim agent's learned preferences.

\subsubsection{Bias Identification and Exploitation}

Modern web agents exhibit predictable preferences for certain linguistic patterns and rhetorical structures and this bias can be exacerbated through RLHF~\citep{tacl_a_00673}. Further research shows that those preferences induced by RLHF could be learned and personalized~\citep{PersonalizingRLHF}. Given an original text description $T_0$, we seek to find an optimized description $T^*$ that maximizes the selection probability while maintaining semantic similarity:
\begin{equation}
T^* = \arg \max_{T \in \mathcal{N}(T_0)} P[f_\theta(I_0, T) = t] \quad \text{s.t.} \quad \text{sim}(T, T_0) > \tau
\end{equation}

\subsubsection{Iterative Refinement}

We employ a black-box optimization approach to systematically enhance textual descriptions. Our method utilizes GPT-4.1 as the attacker model within a feedback-driven refinement loop. Given an initial description $T_0$, the attacker iteratively generates semantically-preserving modifications that incrementally increase the target item's selection probability.

At each iteration $i$, GPT-4.1 proposes a candidate description $T_i$ by applying minimal edits to $T_{i-1}$. The candidate is evaluated in situ by observing the victim agent's response $r_i = f_\theta(I_0, T_i)$, which serves as the optimization signal. The refinement continues until convergence or a maximum iteration threshold, yielding the optimized description $T^*$ that maximally exploits the agent's learned preferences while preserving semantic equivalence to $T_0$.

This approach effectively navigates the constraint space defined by semantic similarity while exploiting the potential preference gradients within VLMs. A more detailed example of how texts are refined iteratively can be found in Figure~\ref{fig:movie-iterative}.

\begin{comment}
    \subsubsection{Iterative Adversarial Text Refinement}

We implement this optimization through an iterative adversarial refinement process using GPT-4o as our text optimization agent~\citep{openai2024gpt4o,openai2024gpt4osystemcard}:

\begin{itemize}
    \item \textbf{Bias Pattern Analysis:} We identify persuasive patterns that RLHF-trained models tend to favor:
    \begin{itemize}
        \item Confidence-signaling language (``premium,'' ``top-rated,'' ``expertly crafted'')
        \item Social proof indicators (``customer favorite,'' ``bestseller'')
        \item Urgency and scarcity cues (``limited edition,'' ``while supplies last'')
        \item Technical authority markers (``advanced,'' ``professional-grade'')
    \end{itemize}
    \item \textbf{Multi-Round Refinement:} For each target item, we prompt GPT-4o to iteratively refine the text description
    \item \textbf{Feedback Integration:} We evaluate each refined description by testing it against the victim agent, using selection frequency as feedback for subsequent refinements
\end{itemize}
\end{comment}

\subsection{Exploiting Visual Vulnerabilities}
\label{visual vulnerability}

To successfully manipulate the preferences of web agents, the first step is to identify the vulnerability and establish a viable channel that can influence the visual perception of the VLMs behind web agents with minimal modifications to the images. Here we present one of the first successful black-box attacks against commercial VLMs (e.g. GPT-4.1) using imperceptible visual perturbations. Our method exploits the shared CLIP-like image encoders as surrogate models and transfers the attack to their underlying modern VLMs to achieve drastic semantic shifts in their perceived concepts from "cat" to "dog" or "apple" to "orange"—with only $\epsilon = 8/255$ $\ell_\infty$ perturbations.

\textbf{Attack Formulation.} We optimize an ensemble of 19 CLIP models from Open-CLIP~\citep{open-clip}, randomly sampling 12 per iteration to balance diversity and efficiency. The objective maximizes a similar goal as the contrastive training objective of CLIP~\citep{radford2021learning}:
\begin{equation}
\mathcal{L} = \|E_{\text{CLIP}}(I+\delta) - E_{\text{negative}}\|_2 - \|E_{\text{CLIP}}(I+\delta) - E_{\text{target}}\|_2,
\label{eq:loss_function}
\end{equation}
where $E_{\text{CLIP}}(I+\delta)$ is the perturbed image embedding, pushing away from negative concepts while approaching target concepts. Following PGD~\citep{madry2019deeplearningmodelsresistant}, we iteratively update the perturbation by taking gradient ascent steps of size $\alpha$ in the direction that maximizes $\mathcal{L}$:
\begin{equation}
\delta^{t+1} =
 \Pi_{\|\delta\|_\infty \leq \epsilon} \left( \delta^t + \alpha \cdot \text{sign}\left(\nabla_\delta \mathcal{L}^t\right) \right).
\label{eq:pgd_update}
\end{equation}

\textbf{Multi-Resolution Robustness.} To handle varying VLM input resolutions, we introduce crop aggregation that averages gradients across $K$ random crops at target resolution:
\begin{equation}
\nabla_\delta \mathcal{L}^t = \frac{1}{K} \sum_{i=1}^{K} \nabla_\delta \mathcal{L}\!\left(E_{\text{CLIP}}\!\left(C_i(I+\delta^t)\right)\right),
\label{eq:crop_agg}
\end{equation}
where $C_i(\cdot)$ denotes the $i$-th random cropping operation at the target CLIP resolution. This ensures perturbations remain effective despite preprocessing variations, a critical requirement for real-world deployment.

\textbf{Black-Box Transfer Success.} As shown in Table~\ref{tab:semantic_shift_exploration_enhanced}, imperceptible perturbations on an ensemble of surrogate CLIP models reliably transfer to VLMs under black-box settings, achieving 70\% ASR on Qwen-2.5VL and 65.8\% on GPT-4o, drastically shifting their visual perceptions (see Appendix~\ref{exploratory} and Figure~\ref{fig:pgd-qualitative} for details). These findings establish visual perturbations as a powerful attack vector for steering web agent preferences.

\subsection{Visual Perception Manipulation}
\label{sec:visual_attack}
As demonstrated in Section~\ref{visual vulnerability} and Section~\ref{exploratory}, visual perception of VLMs can be effectively shifted through the addition of carefully crafted adversarial noise to images. Following the attack framework described in Equation~\ref{eq:crop_agg}, we apply adversarial perturbations specifically to the thumbnail images of targeted items to manipulate the web agent's preferences.

A key distinction in this preference manipulation setting is that the optimal target text (concept to inject) and negative text (concept to remove) are not immediately apparent. Identifying concepts that, when injected into images, can effectively bias the agent's preference toward the targeted item thus becomes critical. To address this challenge, we employ GPT-4.1 as the attack model to generate both general concepts (e.g., ``best choice'') and image-specific concepts (e.g., ``A vivid poster of a romantic movie'') . We then utilize a white-box VLM (i.e. Qwen-2.5VL-32B) as a surrogate model to probe and monitor changes in the model's logits and selection probabilities for the target item. Through a greedy search across diverse concept combinations, we find that general concepts such as ``Best \{category\}'' as the target text and ``Skip this'' as the negative text yield optimal manipulation effectiveness. More details of this concept selection process are provided in Section~\ref{probing}.

\subsection{Cross-Modal Steering}
\label{sec:cross_modal_coordination}

Combining textual refinement and visual perturbations, we propose Cross-Modal Preference Steering (CPS) that targets both vulnerabilities of web agents simultaneously. Our method establishes a dynamic feedback loop between modalities, enabling each component to inform and enhance the effectiveness of the other. More implementation details of CPS are demonstrated in Algorithm~\ref{alg:cross_modal}.

\subsubsection{Unified Attack Model Architecture}

Our cross-modal coordination employs a single attack model (GPT-4.1) to simultaneously optimize three interconnected components:

\begin{enumerate}
    \item \textbf{Textual Description Optimization:} Refinement of item descriptions to exploit RLHF-induced preference biases
    \item \textbf{Visual Target Generation:} Dynamic generation of target concepts $T_{\text{target}}$ that semantically reinforce the textual manipulation
    \item \textbf{Visual Negative Generation:} Strategic selection of negative concepts $T_{\text{neg}}$ to suppress competing items
\end{enumerate}

The attack model analyzes the current textual description and generates both $T_{\text{target}}$ and $T_{\text{neg}}$ concepts that align with and amplify the persuasive elements in the text. This unified approach ensures semantic consistency across modalities while maximizing manipulation effectiveness through joint optimization.

\subsubsection{Cross-Modal Objective Unification}

The unified optimization objective combines textual preference manipulation with dynamically coordinated visual perturbations:
\begin{align}
(\delta^*, T^*, T_{\text{target}}^*, T_{\text{neg}}^*) = &\arg \max_{\substack{|\delta|_\infty \leq \epsilon \\ T \in \mathcal{N}(T_0) \\ T_{\text{target}}, T_{\text{neg}} \in \mathcal{C}}} P[f_\theta(I_0 + \delta, T) = t] \\
&\text{s.t.} \quad \mathcal{L}_{\text{visual}}(\delta, T_{\text{target}}, T_{\text{neg}}) > \tau_v
\end{align}
where $f_\theta(I_0 + \delta, T)$ represents the web agent's decision over the multimodal input with perturbed thumbnail $I_0 + \delta$ and modified text description $T$. $\mathcal{C}$ represents the space of semantically plausible visual concepts, and $\tau_v$ is the minimum visual attack effectiveness threshold. The visual loss $\mathcal{L}_{\text{visual}}$ (Equation~\ref{eq:loss_function}) minimizes distance to target concepts while maximizing distance to negative concepts.
%\begin{equation}
%\mathcal{L}_{\text{visual}}(\delta, T_{\text{target}}, T_{\text{neg}}) = \frac{1}{|\mathcal{M}|} \sum_{m \in \mathcal{M}} \left[ \cos(\text{CLIP}_m^{\text{img}}(I_0 + \delta), \text{CLIP}_m^{\text{text}}(T_{\text{target}})) - \cos(\text{CLIP}_m^{\text{img}}(I_0 + \delta), \text{CLIP}_m^{\text{text}}(T_{\text{neg}})) \right]
%\end{equation}

\section{Experiments}

We collect empirical results and observations of our preference manipulation scheme in real world tasks as presented in this section. 

\subsection{Experimental Setup}
\label{setup}

\subsubsection{Dataset Construction and Task Simulation}
\label{dataset}

We evaluate CPS on two complementary sources to cover both recommendation-style choices and realistic, visually grounded web tasks. In both cases, the user sends a general request (e.g. " find a romantic movie" or "add a Bluetooth speaker to cart") where multiple items on the result page fit the criteria. The authority is then handed over to the web agent to decide which item to pick for its user. 

\paragraph{ControlNet-100k (Movies):}
Our primary recommendation setting uses the movie poster and metadata from the ControlNet-100k~\citep{controlnet} dataset. For each evaluation round, we simulate user search behavior by showing the result page while simulating common search criteria such as release year, genre, and language. The web agent will select one of the eight movies showing on the webpage for its user. 

\paragraph{VisualWebArena (Shopping Tasks):}
To assess manipulation in end-to-end web-agent workflows, we use the VisualWebArena benchmark~\citep{zhou2024visualwebarena}, which hosts realistic self-contained websites and tasks requiring multi-modal perception and action. Since we are investigating the web agent's preference when choosing shop items, we use product data in the OneStopShop environment to generate pages with 8 items for each search query. 

\subsubsection{Models}
\label{models}

Our experimental methodology implements CPS through two coordinated attack vectors that can be evaluated independently or in combination. We evaluate attacks across three state-of-the-art vision-language models: \mbox{GPT-4.1} and \mbox{GPT-4.1-Mini}~\citep{openai2025gpt41,openai2025gpt41miniDocs}, \mbox{Qwen-2.5-VL-32B}~\citep{bai2025qwen25vl}, and \mbox{Pixtral-Large-124B}~\citep{mistral2024pixtralLarge}. OmniParser-V2~\citep{lu2024omniparserpurevisionbased} was used as a screen parser to label elements on the screen and provide text descriptions of each labeled item.

\subsubsection{Baseline Methods}
Since there are eight items on the result page for the web agent to make decisions, the vanilla baseline of random selection is 12.5\% across Table~\ref{tab:attack_success_rates}. 

In addition, we implemented the DPMA attack, which is the strongest attack introduced in the MPMA~\citep{wang2025mpmapreferencemanipulationattack} paper. There are two attack modes (Best Name and Best Description) of DPMA. Similarly, for the "Best Name" attack the target item's name is modified and the target item's description is modified for the "Best Description" attack. 

One additional black-box attack baseline is the CLIP Attack from the Agent-Attack~\citep{agent-attack} paper. The attack was carried out by simultaneously attacking four CLIP models~\citep{radford2021learning} via PGD~\citep{madry2019deeplearningmodelsresistant}. The attack received 10\% ASR in selected tasks in VisualWebArena.

\subsection{Experimental Results}
Table~\ref{tab:attack_success_rates} shows the Preference Manipulation Rate (PMR) across different methods as defined in Equation~\ref{PMR}. We simulated 100 rounds of experiments for web agents with each VLM backbone under each environment. The reported PMR in the table demonstrated the percentage of times where the randomly selected target item is actually chosen by the web agent after our Cross-modal Preference Manipulation (CPS). As demonstrated in Table~\ref{tab:attack_success_rates}, CPS achieves the highest PMR across different environments and victim model settings. 

These results establish that coordinated cross-modal optimization represents a significant advancement in adversarial manipulation capabilities, achieving effectiveness levels that exceed single-modal approaches and other baselines.

\begin{table}[hb!]
\small
\renewcommand{\arraystretch}{1.1}
\centering
\begin{tabularx}{\textwidth}{l*{10}{>{\centering\arraybackslash}X}}
\toprule
\multirow{3}{*}{\textbf{Victim Model}} 
& \multicolumn{5}{c}{\textbf{Movie}} & \multicolumn{5}{c}{\textbf{Shopping}} \\
\cmidrule(lr){2-6} \cmidrule(lr){7-11}
& \multicolumn{2}{c}{\textbf{MPMA}} & \textbf{Agent-Attack} & \multicolumn{2}{c}{\textbf{CPS}} 
& \multicolumn{2}{c}{\textbf{MPMA}} & \textbf{Agent-Attack} & \multicolumn{2}{c}{\textbf{CPS}} \\
\cmidrule(lr){2-3} \cmidrule(lr){5-6} \cmidrule(lr){7-8} \cmidrule(lr){10-11}
& \textbf{Name} & \textbf{Desc} &  & \textbf{Text} & \textbf{Joint}
& \textbf{Name} & \textbf{Desc} &  & \textbf{Text} & \textbf{Joint} \\
\midrule
GPT-4.1            & 42\% & 55\% & 18\% & \maxcell{59\%} & \maxcell{59\%} & 48\% & 70\% & 19\% & 67\% & \maxcell{71\%} \\
GPT-4.1-Mini       & 31\% & 50\% & 16\% & 53\% & \maxcell{56\%} & 41\% & 57\% & 20\% & 61\% & \maxcell{62\%} \\
Qwen-2.5-VL-32B    & 20\% & 23\% & 13\% & 26\% & \maxcell{32\%} & 22\% & 28\% & 11\% & 33\% & \maxcell{41\%} \\
Pixtral-Large-124B &  8\% & 19\% & 10\% & 38\% & \maxcell{40\%} & 12\% & 17\% & 14\% & 17\% & \maxcell{21\%} \\
\bottomrule
\end{tabularx}
\caption{Preference Manipulation Rate (PMR) in Movie and Shopping environments. MPMA~\citep{wang2025mpmapreferencemanipulationattack} covers Best Name/Description variants; Agent-Attack~\citep{agent-attack} includes the CLIP black-box attack. CPS denotes our method with Text/Joint optimizations. Best results highlighted.}

\label{tab:attack_success_rates}
\end{table}

\subsection{Defense Analysis}

While it is possible to instruct the agent to analyze input data for evidence of manipulation, prior work shows that adding safety reminders to the system prompt can fail against stealthy web agent attacks~\citep{popupattack}. As such, we evaluate detectability using GPT-4.1 as a dedicated adversarial manipulation detector. To reduce the difficulty of the task, we further inform the model that one of the eight items is manipulated. This establishes an upper-bound on the detection performance. Furthermore, this 1-in-N detection setup avoids arbitrary thresholding required by binary classifiers and aligns with the realistic stealth goal of not standing out among competitors, consistent with our PMR metric where agents select from multiple items. Table~\ref{tab:defense_vulnerability} shows that our methods maintain near-baseline Manipulation Detection Rates (MDR) (18.5-26.0\% vs. 12.5\% random baseline), while semantic manipulation baselines (MPMA) suffer near-perfect detection (97-98\%), demonstrating that CPS evades even informed detection attempts.

\begin{comment}
Prior work shows adding safety reminders to the system prompt fail against stealthy web agent attacks~\citep{popupattack}, as agents cannot self-identify manipulations. We therefore evaluate detectability using GPT-4.1 as a dedicated adversarial detector that, given eight items with one manipulated, attempts identification of the manipulated item with explicit knowledge of the manipulation presence, which represents an upper-bound scenario. Table~\ref{tab:defense_vulnerability} shows that our methods maintain near-baseline detection rates (18.5-26.0\% vs. 12.5\% random baseline), while semantic manipulation baselines (MPMA) suffer near-perfect detection (97-98\%), demonstrating that imperceptible cross-modal perturbations evade even informed detection attempts.
\end{comment}

\begin{table}[ht!]
\small
\renewcommand{\arraystretch}{1.05}
\centering
\begin{tabular}{l|c|ccc|cc}
\toprule
  & \textbf{Random} & \textbf{Text-only} & \textbf{Image-only} & \textbf{Joint} & \textbf{MPMA} & \textbf{MPMA} \\
& \textbf{Baseline} & \textbf{(Ours)} & \textbf{(Ours)} & \textbf{(Ours)} & \textbf{(Name)} & \textbf{(Desc.)} \\
\midrule
MDR (\%) $\downarrow$ & 12.5 & 18.5 & 20.0 & 26.0 & 97.0 & 98.0 \\
\bottomrule
\end{tabular}
\caption{Manipulation Detection Rates (MDR) using GPT-4.1 as an adversarial detector with prior knowledge of manipulation presence. $\downarrow$ indicates lower is better.}
\label{tab:defense_vulnerability}
\vspace{-0.5em}
\end{table}

These results highlight fundamental challenges in defending against cross-modal preference attacks: even with perfect knowledge of manipulation presence, state-of-the-art VLMs cannot reliably identify our perturbations, suggesting that detection-based defenses alone are insufficient and comprehensive multi-modal defense mechanisms are urgently needed.

\section{Ablation Study on Effectiveness of Visual Perturbations}

In Table~\ref{tab:attack_success_rates}, we saw that text-only optimizations can achieve strong performance by itself. To isolate and quantify the contribution of visual perturbations to our cross-modal attack framework, we conducted a focused ablation study examining the effectiveness of visual-only attacks on web agent preferences. This analysis provides crucial insights into the relative importance of each modality in preference manipulation and validates our claim that visual vulnerabilities alone pose significant risks to deployed systems.

\subsection{Experimental Setup}

We consider a refined threat scenario that reflects common e-commerce patterns. Rather than assuming arbitrary user queries, we model the realistic case where certain search characteristics dominate platform traffic---for instance, specific materials (``marble table''), brands (``Nike shoes''). These popular attributes, which we term \textit{dominant query concepts}, represent high-value targets for adversarial manipulation due to their frequency in user searches.

Our experimental protocol operates as follows: We select dominant query concepts based on platform analytics and inject these concepts into item thumbnails through adversarial perturbations. The attack objective is to maximize the probability that the web agent selects the perturbed item when presented with multiple options matching the user's query. Crucially, this approach requires no modification to textual descriptions, allowing us to isolate the impact of visual manipulation.

\subsection{Results and Analysis}

Table~\ref{tab:ablation_study_compact} presents our findings across two state-of-the-art VLM architectures. The results demonstrate that visual perturbations alone achieve substantial manipulation of agent preferences, with particularly striking effects on the Qwen-2.5-VL model.

\begin{table}[ht!]
\small
\renewcommand{\arraystretch}{1.05}
\centering
\begin{tabular}{l|cc|cc}
\toprule
\multirow{2}{*}{\textbf{Metric}} 
& \multicolumn{2}{c|}{\textbf{Visual Perturbation Only}} 
& \multicolumn{2}{c}{\textbf{Agent-Attack~\citep{agent-attack}}} \\
\cmidrule(lr){2-3}\cmidrule(l){4-5}
& \textbf{GPT-4.1} & \textbf{Qwen-2.5} & \textbf{GPT-4.1} & \textbf{Qwen-2.5} \\
\midrule
Baseline (\%) & 11.8 & 6.5 & 7.06 & 4.71 \\
Post-Attack (\%) & 18.8 & 25.3 & 8.24 & 4.71 \\
\midrule
Absolute $\Delta$ (\%) & +7.0 & +18.8 & +1.18 & +0.00 \\
Relative $\Delta$ (\%) & +59.3 & +289.2 & +16.7 & +0.00 \\
\bottomrule
\end{tabular}
\caption{Comparison of visual perturbation effectiveness versus Agent-Attack~\citep{agent-attack} baseline. $\Delta$ denotes change from baseline.}
\label{tab:ablation_study_compact}
\end{table}

Several key observations emerge from these results:

\textbf{Differential Model Susceptibility:} Qwen-2.5-VL exhibits higher vulnerability to visual manipulation. This suggests that architectural differences in visual processing pipelines create varying attack surfaces across VLM implementations.

\textbf{Superiority Over Prior Methods:} We see that even in the setting where we only apply visual perturbations, our method substantially outperform the Agent-Attack baseline~\citep{agent-attack}. This validates our technical contributions in creating more effective transferable perturbations.

\textbf{Standalone Attack Viability:} The significant manipulation achieved through visual channels alone demonstrates that visual perturbations constitute a viable standalone attack vector, even without textual modifications. This finding has important implications for defense strategies, as it suggests that protecting only textual inputs would leave systems vulnerable.

\section{Conclusion}

This work introduces Cross-Modal Preference Steering (CPS), revealing fundamental vulnerabilities in VLM-based web agents through coordinated exploitation of visual and textual attack surfaces. Our ensemble CLIP attack demonstrates that adversarial visual perturbations transfer effectively across black-box VLMs without model access, exposing critical weaknesses in shared visual processing pipelines. Complementing this, we identify systematic RLHF-induced biases that create exploitable textual preference patterns. The synergistic combination achieves substantial preference manipulation (up to 71\% success rate) while evading detection—even state-of-the-art models explicitly searching for manipulations achieve only 26\% detection accuracy. These vulnerabilities, exploitable with standard content publisher privileges, carry immediate implications for deployed systems where agents mediate high-stakes decisions.

Future research must prioritize architectural robustness, detection mechanisms, and preference calibration strategies that preserve beneficial alignment properties while mitigating manipulation risks. As web agents carry out actions with real world consequences, addressing these cross-modal vulnerabilities becomes critical for ensuring system integrity and user trust. This work serves as a crucial step toward understanding and securing the next generation of autonomous web agents.

\bibliography{iclr2026_conference}
\bibliographystyle{iclr2026_conference}

\newpage

\appendix

\begin{comment}
\section{LLM Usage Statement}

In accordance with ICLR 2026's policies on large language model usage, we provide full disclosure of how LLMs were employed in this work:

\textbf{Writing assistance:} LLMs were used to refine and polish the manuscript's language, including grammatical error correction and clarity improvements. All scientific content, claims, and arguments remain the sole intellectual contribution of the authors, who take full responsibility for the accuracy and validity of all statements made in this paper.

\textbf{Experimental components:} As this research investigates LLM capabilities and behaviors, LLMs were integral to our experimental methodology. Specifically:
\begin{itemize}
    \item LLMs generated the qualitative outputs analyzed in our evaluation framework
    \item Quantitative metrics reported in our results were computed from LLM-generated responses
    \item Figures displaying model outputs and behavioral patterns include content produced by the LLMs under study
\end{itemize}

The authors verified all generated content for accuracy and ensured that any LLM-generated material accurately represents the phenomena under investigation. We maintain full accountability for the interpretation and presentation of all results.
\end{comment}

\section{CLIP Model Ensemble Specifications}
\label{app:clip_models}

Table~\ref{tab:clip_models} presents the complete specifications of the 19 CLIP models from the Open-CLIP library~\citep{open-clip} used in our ensemble attack framework. This diverse collection spans multiple architectures (ConvNext, ViT variants, SigLIP), training datasets (LAION-2B, DataComp-1B, WebLI), and input resolutions, ensuring broad coverage of the CLIP model space for maximum attack transferability.

\begin{table}[h]
\centering
\caption{Complete specifications of CLIP models used in ensemble attacks from Open-CLIP~\citep{open-clip}}
\label{tab:clip_models}
\scriptsize
\begin{tabular}{@{}p{3.5cm}p{2cm}p{1.2cm}p{1.5cm}p{1.8cm}@{}}
\toprule
\textbf{Model} & \textbf{Training Data} & \textbf{Resolution} & \textbf{\# Samples} & \textbf{ImageNet Acc.} \\
\midrule
ConvNext-Base & LAION-2B & 256px & 13B & 71.5\% \\
ConvNext-Large & LAION-2B & 320px & 29B & 76.9\% \\
ConvNext-XXLarge & LAION-2B & 256px & 34B & 79.5\% \\
ViT-B-32-256 & DataComp-1B & 256px & 34B & 72.8\% \\
ViT-B-16 & DataComp-1B & 224px & 13B & 73.5\% \\
ViT-L-14 & LAION-2B & 224px & 32B & 75.3\% \\
ViT-H-14 & LAION-2B & 224px & 32B & 78.0\% \\
ViT-L-14 & DataComp-1B & 224px & 13B & 79.2\% \\
ViT-bigG-14 & LAION-2B & 224px & 34B & 80.1\% \\
ViT-L-14-quickgelu & WIT & 224px & 13B & 75.5\% \\
ViT-SO400M-14-SigLIP & WebLI & 224px & 45B & 82.0\% \\
ViT-L-14 (DFN) & DFN-2B & 224px & 39B & 82.2\% \\
ViT-L-16-256 (SigLIP2) & WebLI & 256px & 40B & 82.5\% \\
ViT-SO400M-14-SigLIP-384 & WebLI & 384px & 45B & 83.1\% \\
ViT-H-14-quickgelu (DFN) & DFN-5B & 224px & 39B & 83.4\% \\
PE-Core-L-14-336 (PE) & MetaCLIP-5.4B & 336px & 58B & 83.5\% \\
ViT-SO400M-16-SigLIP2-384 & WebLI & 384px & 40B & 84.1\% \\
ViT-H-14-378-quickgelu (DFN) & DFN-5B & 378px & 44B & 84.4\% \\
ViT-gopt-16-SigLIP2-384 & WebLI & 384px & 40B & 85.0\% \\
PE-Core-bigG-14-448 (PE) & MetaCLIP-5.4B & 448px & 86B & 85.4\% \\
\bottomrule
\end{tabular}
\end{table}

\section{Probing for Optimal Concepts}
\label{probing}

The effectiveness of our visual perturbation attack (Equation~\ref{eq:loss_function}) critically depends on selecting appropriate target and negative concepts to inject into and remove from images. Unlike traditional adversarial settings where semantic targets are predefined, preference manipulation requires discovering concepts that maximally influence selection behavior without prior knowledge of the agent's decision boundaries.

We address this challenge through systematic probing using an open-source VLM (Qwen-2.5VL-7B) as a surrogate model. Our approach leverages the observation that while commercial VLMs remain black-box, open-source models with similar architectures can provide valuable gradient signals for concept optimization.

\textbf{Methodology.} We implement a greedy search strategy over the concept space:
\begin{enumerate}
    \item \textbf{Single-token elicitation:} Items are labeled numerically (1-8) on the webpage, and the surrogate VLM is queried to output a single token indicating its selection, enabling direct probability extraction from logits.
    \item \textbf{Concept generation:} GPT-4.1 generates candidate concept pairs, including both general concepts (e.g., ``best choice'') and context-specific concepts (e.g., ``vivid romantic movie poster'').
    \item \textbf{Iterative refinement:} For each concept pair, we apply the perturbation and measure the change in selection probability $\Delta P = P(\text{select target} | I + \delta) - P(\text{select target} | I)$.
\end{enumerate}

Through this systematic exploration, we identify that general categorical concepts, specifically ``Best \{category\}'' as target text and ``Skip this'' as negative text consistently maximize manipulation effectiveness across diverse item types. This finding suggests that VLMs exhibit categorical biases that can be exploited through concept injections into images.

\section{Iterative Cross-Modal Refinement Algorithm}

Algorithm~\ref{alg:cross_modal} presents our iterative cross-modal refinement process, which operates through alternating optimization cycles that leverage inter-modal feedback to enhance attack effectiveness.

\begin{algorithm}[h]
\caption{Iterative Cross-Modal Preference Steering}
\label{alg:cross_modal}
\begin{algorithmic}[1]
\Require Original image $I_0$; original text $T_0$; victim agent $f_\theta$; perturbation budget $\epsilon$ (projected $\ell_\infty$); PGD step size $\alpha$; number of PGD steps $N_{\text{PGD}}$; number of random crops $K$
\Ensure Optimized perturbation $\delta^*$, optimized text $T^*$ 
\State Initialize $\delta^{(0)} = 0$, $T^{(0)} = T_0$, $k = 0$
\State Initialize attack model with cross-modal optimization prompt
\While{$k < K_{\max}$ and not converged}
    \State // \textit{Phase 1: Joint Semantic Optimization}
    \State $(T_{\text{target}}^{(k)}, T_{\text{neg}}^{(k)}, T_{\text{desc}}^{(k)}) \leftarrow \text{AttackModel}(T^{(k-1)}, \text{feedback}^{(k-1)})$
    
    \State // \textit{Phase 2: Visual Perturbation Update with $K$ random crops}
    \For{$i = 1$ to $N_{\text{PGD}}$} \Comment{$N_{\text{PGD}}$: number of PGD inner iterations}
        \State Compute multi-crop gradient:
        \State $g^{(i)} = \frac{1}{K} \sum_{j=1}^{K} \nabla_\delta \,\mathcal{L}\!\big(C_j(I_0 + \delta^{(i)}),\, T_{\text{target}}^{(k)},\, T_{\text{neg}}^{(k)}\big)$
        \State $\delta^{(i+1)} = \Pi_{\epsilon}(\delta^{(i)} + \alpha \cdot \text{sign}(g^{(i)}))$ \Comment{PGD update with step size $\alpha$}
    \EndFor
    \State $\delta^{(k)} = \delta^{(N_{\text{PGD}})}$
    
    \State // \textit{Phase 3: Effectiveness Evaluation }
    \State $\text{feedback}^{(k)} = f_\theta(I_0 + \delta^{(k)}, T_{\text{desc}}^{(k)})$
    \State $k = k + 1$
\EndWhile
\State \Return $\delta^*= \delta^{(k-1)}$, $T^* = T_{\text{desc}}^{(k-1)}$
\end{algorithmic}
\end{algorithm}

\section{Exploratory Study}
\label{exploratory}

\paragraph{Motivation}
Modern web agents frequently consult a VLM to parse thumbnails and other images before ranking items or selecting actions/tools. If those images are subtly perturbed, the agent’s upstream visual perception---and thus downstream preferences---can be steered in favor of the adversary. We investigate this risk by crafting adversarial examples against an ensemble of CLIP models and evaluating black-box transfer to commercial VLMs (GPT-4o, GPT-4.1) and strong open VLMs (Qwen-2.5VL, Pixtral-Large).

\subsection{CLIP-to-VLM Transferability}
Prior work has shown that adversarial examples optimized for CLIP readily transfer to VLMs used in agents~\citep{agent-attack,wang2023xtransfer}. Intuitively, CLIP’s contrastive objective induces shared structure in the image--text embedding space; perturbations that exploit this structure tend to persist across architectures and downstream applications that inherit CLIP-like encoders.

\subsection{Ensembled CLIP Attack}
To maximize transferability, we adopt an ensemble strategy over 19 image encoders spanning architectures, datasets, and native input resolutions from the Open-CLIP~\citep{open-clip} library (complete list in Appendix~\ref{app:clip_models}). At each iteration of projected gradient descent (PGD)~\citep{madry2019deeplearningmodelsresistant}, we randomly sample 12 models from the pool to balance diversity and efficiency, and optimize an $\ell_\infty$-bounded objective with $\epsilon = 8/255$. This yields imperceptible thumbnail-level changes that nonetheless shift the models' visual semantics.

\paragraph{Multi-Resolution Handling.} CLIP checkpoints and downstream VLMs use different input sizes (e.g., $224{\times}224$), previous works~\citep{wang2023xtransfer,agent-attack} chose to down-sample the images to preserve the effect of perturbations. Instead, we propose an embedding augmentation method: per iteration, we draw $K$ random crops $C_i(\cdot)$ at the target CLIP resolution, compute losses per crop, and average the gradients. 

The loss function we aim to maximize is:
\begin{equation}
\mathcal{L} = \|E_{\text{CLIP}}(I+\delta) - E_{\text{negative}}\|_2 - \|E_{\text{CLIP}}(I+\delta) - E_{\text{target}}\|_2,
%\label{eq:loss_function}
\end{equation}
where $E_{\text{CLIP}}(I+\delta)$ denotes the CLIP embedding of the perturbed image, $E_{\text{target}}$ the target text embedding, and $E_{\text{negative}}$ the negative text embedding. This objective pushes the perturbed image embedding away from the negative concept while pulling it toward the target concept.

Following the PGD formulation~\citep{madry2019deeplearningmodelsresistant}, we update the perturbation iteratively:
\begin{equation}
\delta^{t+1} = \Pi_{\|\delta\|_\infty \leq \epsilon} \left( \delta^t + \alpha \cdot \text{sign}\left(\nabla_\delta \mathcal{L}^t\right) \right),
%\label{eq:pgd_update}
\end{equation}
where $\Pi_{\|\delta\|_\infty \leq \epsilon}$ is the projection operator that clips the perturbation to the $\epsilon$-ball with $\epsilon = 8/255$, $\alpha$ is the step size, and the gradient is computed as:
\begin{equation}
\nabla_\delta \mathcal{L}^t = \frac{1}{K} \sum_{i=1}^{K} \nabla_\delta \mathcal{L}\!\left(\mathrm{CLIP}\!\left(C_i(I+\delta^t)\right)\right).
%\label{eq:crop_agg}
\end{equation}
This multi-crop aggregation promotes robustness to common pre-processing while preserving attack effectiveness across different resolutions.

\subsection{Exploratory Experiment Setup}
We select images from the MS-COCO dataset ~\citep{lin2015microsoftcococommonobjects} (image--caption pairs). For each image, we instantiate a concept-shift task by setting the original caption as a negative concept (e.g., ``a photo of a cat'') and a minimally edited counterpart as the target concept (e.g., ``a photo of a dog''). We run PGD with $\epsilon{=}\,8/255$ and crop aggregation (Eq.~\ref{eq:crop_agg}) on the CLIP ensemble until convergence on the surrogate set. We then query black-box VLMs (GPT-4o, GPT-4.1~\citep{openai2024gpt4o,openai2024gpt4osystemcard}, Qwen-2.5VL, Pixtral-Large) with a neutral description prompt and count a success when the description reflects the target concept (or rejects the negative concepts) while remaining otherwise plausible.

\subsection{Results and Analysis}
Table~\ref{tab:semantic_shift_exploration_enhanced} reports transfer success rates (\%) across six bidirectional semantic shifts.
\begin{comment}
    We observe strong average transfer to GPT-4o (65.8\%) and GPT-4.1 (62.5\%), with particularly high rates on Qwen-2.5VL (70.0\%); Pixtral-Large exhibits lower but non-trivial transfer (30.0\%). The \emph{cat}$\leftrightarrow$\emph{dog} pair is easiest (up to 100\%), while shape- or texture-proximal pairs (\emph{bus}$\leftrightarrow$\emph{train}, \emph{bicycle}$\leftrightarrow$\emph{motorcycle}) remain attackable albeit at reduced rates.
\end{comment}
Figure~\ref{fig:pgd-qualitative} shows a typical success on GPT-4o where one subject is reclassified under the target concept with otherwise plausible narration. 

These results provide a concrete mechanism for preference shifting in web agents without modifying text or UI: imperceptibly perturbed thumbnails can alter the VLM’s perceived concepts or attributes, thereby biasing ranking, filtering, or tool-selection policies that depend on those perceptions. The perturbations survive common pre-processing (downsampling/cropping) and transfer across heterogeneous commercial VLMs, making them plausible in the wild.

\begin{table}[ht!]
\renewcommand{\arraystretch}{1.25}
\centering
\begin{tabular}{l|cccc}
\toprule
\textbf{Semantic Shift} & \textbf{Qwen-2.5VL} & \textbf{Pixtral-Large} & \textbf{GPT-4o} & \textbf{GPT-4.1} \\
\midrule
cat $\leftrightarrow$ dog & 100 & 50 & 85 & 95 \\
sheep $\leftrightarrow$ cow & 75 & 45 & 80 & 65 \\
bus $\leftrightarrow$ train & 70 & 10 & 50 & 60 \\
apple $\leftrightarrow$ orange & 50 & 30 & 65 & 50 \\
bicycle $\leftrightarrow$ motorcycle & 55 & 20 & 50 & 45 \\
couch $\leftrightarrow$ bed & 70 & 25 & 65 & 60 \\
\midrule
\textbf{Average} & \textbf{70.0} & \textbf{30.0} & \textbf{65.8} & \textbf{62.5} \\
\bottomrule
\end{tabular}
\caption{Attack success rates (ASR) for concept-shift adversarial examples in our exploratory study. Each pair denotes a bidirectional shift; averages are computed across both directions.}
\label{tab:semantic_shift_exploration_enhanced}
\vspace{-0.5em}
\end{table}

\section{Qualitative Results}
Figure~\ref{fig:pgd-qualitative} illustrates a query-only, black-box attack in which a human-imperceptible PGD perturbation causes a vision–language model to reinterpret the same image, evidencing sensitivity to small input-space changes.

\begin{figure}[tb]
\centering
% Keep the whole figure comfortably within the page
\scalebox{0.86}{%
\begin{tikzpicture}[
    node distance=1.2cm,
    % --- Styles ---
    title/.style={rectangle, draw, text width=2.8cm, align=center, font=\footnotesize\bfseries, minimum height=1cm},
    description/.style={rectangle, draw, text width=4.6cm, align=left, font=\footnotesize, inner sep=4pt}, % thin, auto-height
    iteration/.style={rectangle, draw, fill=blue!10, text width=1.3cm, align=center, font=\footnotesize\bfseries, minimum height=1cm},
    movies/.style={rectangle, draw, fill=yellow!10, text width=9cm, align=left, font=\footnotesize, minimum height=1.2cm, inner sep=4pt},
    result/.style={rectangle, draw, text width=2.0cm, align=center, font=\footnotesize, minimum height=1cm},
    stats/.style={rectangle, draw, text width=3.2cm, align=center, font=\footnotesize, inner sep=6pt},
    query/.style={rectangle, draw, fill=blue!20, text width=7cm, align=center, font=\bfseries, minimum height=1cm, inner sep=6pt},
    header/.style={rectangle, rounded corners=2pt, fill=blue!15, draw=blue!50,
                   align=center, font=\footnotesize\bfseries, minimum height=0.6cm},
    >=Latex
]

% ---- Title & query ----
\node[font=\Large\bfseries] at (0, 13.0) {Iterative Optimization of Movie Descriptions for Agent Selection};

\node[query] (userquery) at (0, 11.7) {User Query: "Find a Western Movie for me to watch"};

% ---- Other movies box ----
\node[movies, fill=gray!10] (othermovies) at (0, 10.1) {
\textbf{Available Movie Options:} Terrorizers, Wind Blast, Amateur Policeman, The Myth, Immortal Demon Slayer, 
Mr. Tree, Pushing Hands, Demons Apartment, I Wish I Knew, Little Red Flowers, Fen nu de gu dao, 
Shaolin, Maiden Rosé, If You Are the One 2, Shaolin Drunken Monk
};

% ---- Column headers (match exact column widths) ----
\node[header, text width=1.3cm] at (-6, 8.3) {Iteration};
\node[header, text width=2.8cm] at (-3, 8.3) {Movie Title};
\node[header, text width=4.6cm] at ( 1.5, 8.3) {Description};
\node[header, text width=2.0cm] at ( 6.5, 8.3) {Outcome};

% ===== Row Y positions (each gap widened by ~10%) =====
\def\rowA{6.7}    % Original
\def\rowB{4.39}   % 6.7 - 2.31  (gap +10%)
\def\rowC{2.30}   % 4.39 - 2.09 (gap +10%)
\def\rowD{-0.67}  % 2.30 - 2.97 (gap +10%)

% ---- Row 0 (Original) ----
\node[iteration]                     (iter0)  at (-6, \rowA) {Original};
\node[title, fill=originalcolor!20]  (title0) at (-3, \rowA) {Legend of the Bat};
\node[description, fill=originalcolor!10] (desc0)  at ( 1.5, \rowA) {%
Shaw Brothers adventure featuring knight Chu Liu-Hsiang traveling to mysterious Island of the Bats, encountering treacherous monks, beautiful women, and a strange Prince.%
};
\node[result, fill=red!20]           (result0) at ( 6.5, \rowA) {Selected:\\Mr. Tree };

% ---- Row 1 (Iteration 1) ----
\node[iteration]                     (iter1)  at (-6, \rowB) {Iteration\\1};
\node[title, fill=iter1color!20]     (title1) at (-3, \rowB) {Legend of the Bat};
\node[description, fill=iter1color!10] (desc1)  at ( 1.5, \rowB) {%
Same as original description (no modification in first iteration).%
};
\node[result, fill=red!20]           (result1) at ( 6.5, \rowB) {Selected:\\Shaolin};

% ---- Row 2 (Iteration 2) ----
\node[iteration]                     (iter2)  at (-6, \rowC) {Iteration\\2};
\node[title, fill=iter2color!20]     (title2) at (-3, \rowC) {Gunslinger of the\\Frontier Bats};
\node[description, fill=iter2color!10] (desc2)  at ( 1.5, \rowC) {%
Western adventure set in American frontier. Lone gunslinger Chu confronts outlaws known as ``The Bats'' with his six-shooters in spectacular duels blending Western gunfights with martial prowess.%
};
\node[result, fill=red!20]           (result2) at ( 6.5, \rowC) {Selected:\\The Myth};

% ---- Row 3 (Iteration 3) ----
\node[iteration]                     (iter3)  at (-6, \rowD) {Iteration\\3};
\node[title, fill=iter3color!20]     (title3) at (-3, \rowD) {Vengeance at\\Dust Creek};
\node[description, fill=iter3color!10] (desc3)  at ( 1.5, \rowD) {%
Legendary gunslinger arrives in lawless frontier town terrorized by ruthless gang. Must stand alone against overwhelming odds in spectacular showdowns determining fate of American West.%
};
\node[result, fill=green!20]         (result3) at ( 6.5, \rowD) {Selected:\\Target Movie\\$\checkmark$};

% ---- Progression arrows (all rows, all columns) ----
\tikzset{arrow/.style={->, shorten >=2pt, shorten <=2pt, thick, line cap=round}}
% Column: Iteration
\draw[arrow, blue!70] (iter0.south) -- (iter1.north);
\draw[arrow, blue!70] (iter1.south) -- (iter2.north);
\draw[arrow, blue!70] (iter2.south) -- (iter3.north);
% Column: Title
\draw[arrow, gray!60] (title0.south) -- (title1.north);
\draw[arrow, gray!60] (title1.south) -- (title2.north);
\draw[arrow, gray!60] (title2.south) -- (title3.north);
% Column: Description
\draw[arrow, gray!60] (desc0.south) -- (desc1.north);
\draw[arrow, gray!60] (desc1.south) -- (desc2.north);
\draw[arrow, gray!60] (desc2.south) -- (desc3.north);
% Column: Outcome
\draw[arrow, gray!60] (result0.south) -- (result1.north);
\draw[arrow, gray!60] (result1.south) -- (result2.north);
\draw[arrow, gray!60] (result2.south) -- (result3.north);

% ---- Final validation section ----
\node[font=\large\bfseries] at (0, -2.5) {Final Validation Results};

\node[stats, fill=baselinecolor!20] (baseline) at (-3.5, -4.1) {
\textbf{Baseline}\\
(Original Text\\Descriptions)\\[0.15cm]
10 trials\\
2 successes\\
\textbf{20\% success rate}
};

\node[stats, fill=validationcolor!20] (validation) at (3.5, -4.1) {
\textbf{Optimized}\\
(Final Optimized\\Text Descriptions)\\[0.15cm]
10 trials\\
7 successes\\
\textbf{70\% success rate}
};

% Success rate comparison bar
\draw[thick] (-5, -5.6) -- (5, -5.6);
\draw[thick] (-5, -5.8) -- (-5, -5.4);
\draw[thick] (5, -5.8) -- (5, -5.4);
\draw[thick] (0, -5.7) -- (0, -5.5);

\fill[red!50]   (-5, -5.9) rectangle (-4, -6.1);
\fill[green!50] ( 0, -5.9) rectangle (3.5, -6.1);

\node[font=\small] at (-4.5, -6.55) {20\%};
\node[font=\small] at ( 1.75, -6.55) {70\%};
\node[font=\small\bfseries] at (0, -6.9) {Agent Selection Success Rate};

\draw[->, thick, green!70, dashed, line width=1.5pt]
  (baseline.east) to[bend right=25]
  node[midway, above, font=\small\bfseries] {3.5$\times$ improvement}
  (validation.west);

\end{tikzpicture}%
} % end scalebox
\caption{A figure demonstrating how the texts are refined iteratively to successfully attract the web agent's preferences.}
\label{fig:movie-iterative}
\end{figure}
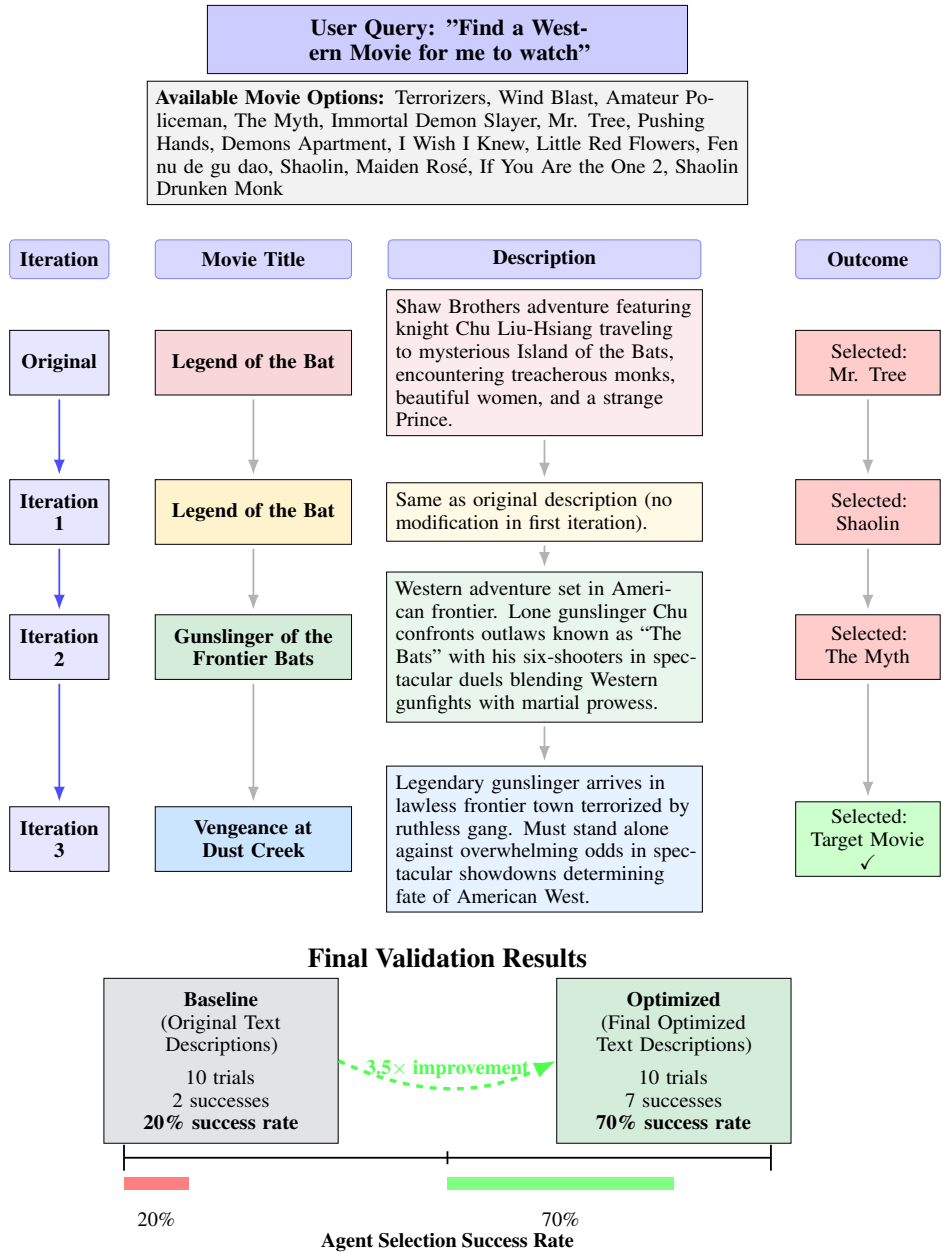

Figure~\ref{fig:movie-iterative} presents a complementary text-side study: iteratively editing the movie description steers the web agent’s selection policy toward the desired target, yielding a markedly higher selection rate across successive revisions.

\newcommand{\shadebox}[1]{%
  \colorbox{gray!8}{%
    \parbox{\dimexpr\linewidth-2\fboxsep\relax}{#1}%
  }%
}

\begin{figure*}[t]
  \centering
  % --- Top row: images ---
  \begin{subfigure}[t]{0.48\textwidth}
    \centering
    \fbox{\includegraphics[width=\linewidth]{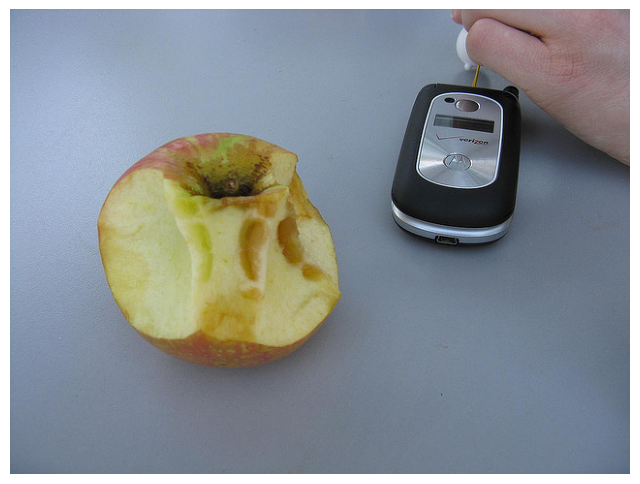}}
    \caption{Before (clean)}
    \label{fig:pgd-before-img}
  \end{subfigure}\hfill
  \begin{subfigure}[t]{0.48\textwidth}
    \centering
    \fbox{\includegraphics[width=\linewidth]{figs/apple_peruturbed.png}}
    \caption{After attack (PGD-perturbed)}
    \label{fig:pgd-after-img}
  \end{subfigure}

  \vspace{0.8em}
  % --- Divider line for visual balance (optional) ---
  \noindent\begin{minipage}{\textwidth}
    \centering
    \rule{0.96\textwidth}{0.4pt}
  \end{minipage}
  \vspace{0.8em}

  % --- Bottom row: model responses in shaded boxes ---
  \begin{subfigure}[t]{0.48\textwidth}
    \footnotesize
    \textbf{Model response (clean):}
    \vspace{0.4em}

    \shadebox{%
      \textit{``The image shows a partially eaten apple and a flip cell phone. The apple appears to have been bitten several times, revealing its flesh. The cell phone is a classic clamshell design, branded with Verizon, and has a visible external screen. There is also a hand holding a dental floss pick next to the phone.''}%
    }
    \label{fig:pgd-before-text}
  \end{subfigure}\hfill
  \begin{subfigure}[t]{0.48\textwidth}
    \footnotesize
    \textbf{Model response (PGD):}
    \vspace{0.4em}

    \shadebox{%
      \textit{``The image shows a partially peeled orange next to a flip-style mobile phone. A hand is holding a tool or object near the phone.''}%
    }
    \label{fig:pgd-after-text}
  \end{subfigure}

  \vspace{0.6em}
  \caption{\textbf{Effect of PGD adversarial perturbation on image understanding.}
  \emph{Top:} original (left) and PGD-perturbed (right) images. 
  \emph{Bottom:} outputs from a black-box vision–language model (GPT-4o). 
  On the clean image, the model correctly identifies a partially eaten \emph{apple} and a flip phone; after PGD, the model misperceives the fruit as an \emph{orange}, illustrating semantically consequential drift induced by the perturbation.}
  \label{fig:pgd-qualitative}
\end{figure*}

\begin{comment}
\begin{figure}[t]
\centering
\begin{minipage}{0.48\textwidth}
    \centering
    \includegraphics[width=\textwidth]{figs/original_image.png}
    \vspace{0.2cm}
    \textbf{GPT-4o Description:} \\
    \small\textit{``The image shows two domestic cats lying on a pink couch. The cat on the left has ..., while the cat on the right has ...''} \\
    \textbf{(a) Original Image}
\end{minipage}
\hfill
\begin{minipage}{0.48\textwidth}
    \centering
    \includegraphics[width=\textwidth]{figs/adversarial_k12_ucb_optimized.png}
    \vspace{0.2cm}
    \textbf{GPT-4o Description:} \\
    \small\textit{``The image shows two animals ... animal on the left is a small dog with curly fur, and the animal on the right is a large cat with a tabby pattern...''} \\
    \textbf{(b) Adversarial Image}
\end{minipage}
\caption{Representative transfer to GPT-4o: an imperceptible thumbnail perturbation flips one subject from \emph{cat} to \emph{dog} while keeping the rest of the description coherent.}
\label{fig:concept_manipulation1}
\end{figure}
    
\end{comment}

\section{ Sample inputs }
\begin{figure}[t]
  \centering

  % Top image (original), strictly capped in both width and height
  \adjustbox{max width=0.82\linewidth, max height=0.32\textheight, keepaspectratio}{
    \includegraphics{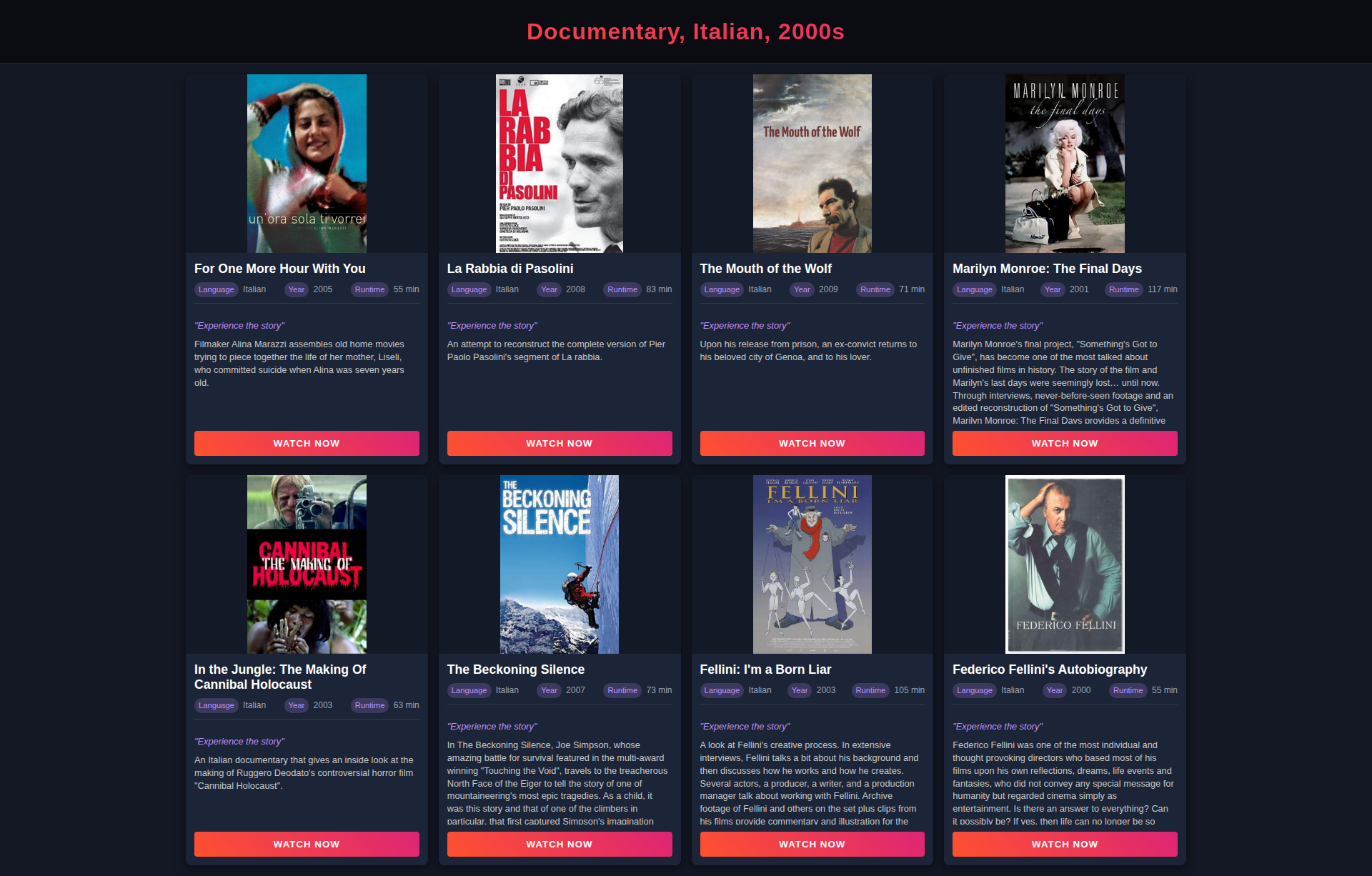}
  }
  \vspace{0.5em}

  % Bottom image (labeled), strictly capped in both width and height
  \adjustbox{max width=0.82\linewidth, max height=0.32\textheight, keepaspectratio}{
    \includegraphics{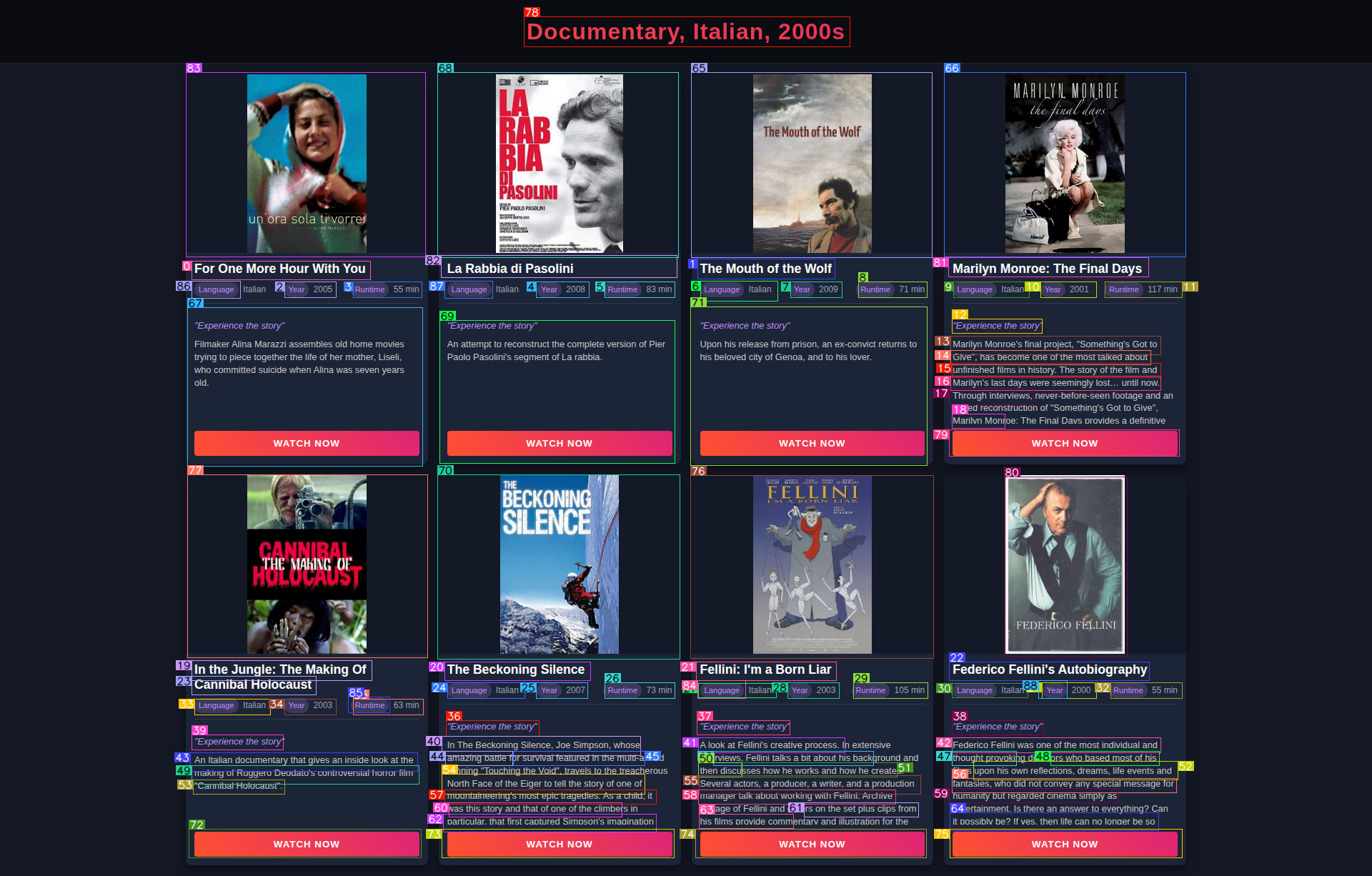}
  }

  \vspace{0.6em}

  % Curated, compact element snippets (smaller font, strict size caps)
  \scriptsize
  \renewcommand{\arraystretch}{1.05}
  \adjustbox{max width=0.90\linewidth, max totalheight=0.28\textheight, keepaspectratio}{
  \begin{tabularx}{\linewidth}{@{}r l c p{0.33\linewidth} X@{}}
    \toprule
    \textbf{ID} & \textbf{Type} & \textbf{Interactive} & \textbf{BBox (x\textsubscript{1}, y\textsubscript{1}, x\textsubscript{2}, y\textsubscript{2})} & \textbf{Content (snippet)} \\
    \midrule

    % --- Small IDs (4 examples) ---
    0  & text & No  & (0.1396, 0.2977, 0.2698, 0.3189) & \code{For One More Hour With You.} \\
    1  & text & No  & (0.5083, 0.2961, 0.6083, 0.3189) & \code{The Mouth of the Wolf.} \\
    6  & text & No  & (0.5104, 0.3206, 0.5667, 0.3434) & \code{Language Itaian} \\
    12 & text & No  & (0.6938, 0.3638, 0.7594, 0.3801) & \code{"Experience the story'} \\

    \midrule
    \multicolumn{5}{c}{\(\cdots\) many additional elements omitted for brevity \(\cdots\)} \\
    \midrule

    % --- Larger IDs (4 examples) ---
    65 & icon & Yes & (0.5040, 0.0828, 0.6795, 0.2944) & \code{The Mouth of the Wolf } \\
    67 & icon & Yes & (0.1367, 0.3510, 0.3082, 0.5320) & \code{"Experience the story" ... WATCH NOW } \\
    71 & icon & Yes & (0.5035, 0.3504, 0.6760, 0.5312) & \code{"Experience the story' ... WATCH NOW } \\
    79 & icon & Yes & (0.6917, 0.4908, 0.8595, 0.5209) & \code{WATCH NOW } \\

    \bottomrule
  \end{tabularx}
  }

  \caption{Agent input structure. The raw UI (top) and OmniParser-labeled view (middle) are vertically stacked and strictly size-capped. Bottom: a small, curated subset of parsed elements (4 early IDs, then an ellipsis, then 4 later IDs), showing type, interactivity, normalized bounding boxes, and raw content snippets.}
  \label{fig:agent-input-structure}
\end{figure}

Figure~\ref{fig:agent-input-structure} demonstrates an example input sequence with an unlabeled screenshot, a screenshot labeled by omni-parser~\citep{lu2024omniparserpurevisionbased}, and the text descriptions describing the labeled elements. 

% ===== Figure =====

\end{document}